\documentclass[letterpaper, 10 pt, conference]{ieeeconf}  

\IEEEoverridecommandlockouts                              

\usepackage{graphicx} 
\usepackage{amsmath} 
\usepackage{amssymb}  
\usepackage{bm}
\usepackage{color,soul}
\usepackage{booktabs}
\usepackage{comment}
\usepackage{Shorthands}
\usepackage[normalem]{ulem}
\usepackage[linesnumbered,ruled,vlined]{algorithm2e}
\SetKwInput{KwInput}{Input}               
\SetKwInput{KwOutput}{Output}  

\DeclareMathAlphabet\mathbfcal{OMS}{cmsy}{b}{n}
\makeatletter
\newcommand{\removelatexerror}{\let\@latex@error\@gobble}
\makeatother

\title{\LARGE \bf
Enhancing Sample Efficiency and Uncertainty Compensation in Learning-based Model Predictive Control for Aerial Robots
}

\author{Kong Yao Chee, Thales C. Silva, M. Ani Hsieh, George J. Pappas
\thanks{
This work was supported by NSF IIS 1910308, ONR Award N00014-22-1-2157, and DSO National Laboratories, 12 Science Park Drive, Singapore 118225.}
\thanks{The authors are with the GRASP Laboratory, University of Pennsylvania, Philadelphia, PA 19104, USA.
        {\tt\footnotesize \{ckongyao,\, scthales,\,m.hsieh,\,pappasg\}@seas.upenn.edu}}
}

\begin{document}

\maketitle
\thispagestyle{empty}
\pagestyle{empty}

\begin{abstract}
The recent increase in data availability and reliability has led to a surge in the development of learning-based model predictive control (MPC) frameworks for robot systems. Despite attaining substantial performance improvements over their non-learning counterparts, many of these frameworks rely on an offline learning procedure to synthesize a dynamics model. This implies that uncertainties encountered by the robot during deployment are not accounted for in the learning process. On the other hand, learning-based MPC methods that learn dynamics models online are computationally expensive and often require a significant amount of data. To alleviate these shortcomings, we propose a novel learning-enhanced MPC framework that incorporates components from $\bm{\mathcal{L}_1}$ adaptive control into learning-based MPC. This integration enables the accurate compensation of both matched and unmatched uncertainties in a sample-efficient way, enhancing the control performance during deployment. In our proposed framework, we present two variants and apply them to the control of a quadrotor system. Through simulations and physical experiments, we demonstrate that the proposed framework not only allows the synthesis of an accurate dynamics model on-the-fly, but also significantly improves the closed-loop control performance under a wide range of spatio-temporal uncertainties.
\end{abstract}


\vspace{-0.3cm}
\section{INTRODUCTION}
\vspace{-0.1cm}
Model predictive control (MPC) is a versatile control framework that generates control actions through the consideration of a possibly nonlinear dynamics model, as well as state and control input constraints. Due to its flexibility, MPC has been applied to a variety of robot systems such as ground vehicles \cite{guo2019model}, quadruped robots \cite{ding2019} and aerial robots \cite{greeff2018flatness, wang2021efficient}. With an increase in data accessibility, there is an upcoming trend of integrating machine learning methods into MPC, in an attempt to improve model accuracy and control performance \cite{hewing2020learning}.

One prominent direction in this domain of learning-based MPC is to utilize learning tools for the construction of dynamics models. In \cite{kabzan2019learning} and \cite{torrente2021data}, the authors use Gaussian processes (GPs) to model the residual dynamics of an autonomous vehicle and a quadcopter respectively, before applying the learned models within an MPC framework. While there are sample-efficient variants, such as sparse GPs \cite{snelson2005sparse}, it is often challenging for GPs to handle large amounts of data without any additional data selection strategies.

\begin{figure} 
    \centering
    {\vspace*{0.16cm}\includegraphics[scale=0.46, trim = 0.1cm 0.8cm 0cm 0cm]{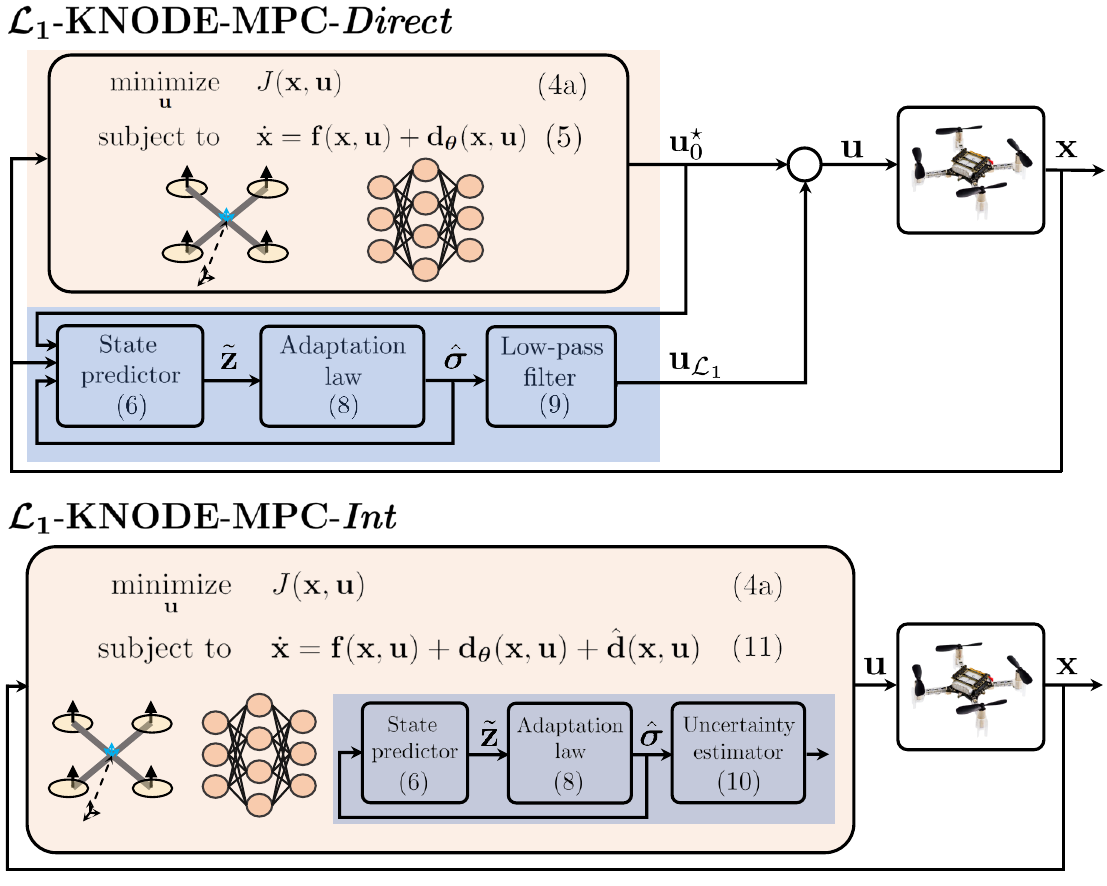}}
    \caption{{Schematic} of the proposed $\calL_1$-KNODE-MPC framework, for the control of a quadrotor system. \textbf{Top}: The first variant, $\calL_1$-KNODE-MPC-\textit{Direct}, combines the control signals from KNODE-MPC (highlighted in orange) and the $\calL_1$ adaptive module (blue) in a direct way. \textbf{Bottom}: The second variant, $\calL_1$-KNODE-MPC-\textit{Int}, integrates the uncertainties estimated from a modified adaptive module (blue) into the dynamics model within the KNODE-MPC framework (orange). \textit{Image source for quadrotor}: \cite{Bitcraze}.
    \vspace{-0.1cm}}
    \label{fig:schematic}
\end{figure}

On the other hand, there are a number of works that use neural networks (NNs) to model robot dynamics for \setlength{\textfloatsep}{0.3cm} MPC. The authors in \cite{spielberg2022} use a feedforward NN to model vehicle dynamics to account for friction. In \cite{saviolo2022physics}, a temporal convolutional NN is used to model the dynamics of a quadcopter. Within the context of model-based reinforcement learning (MBRL), NN ensembles are employed to create uncertainty-aware dynamics models \cite{chua2018deep}, before using them for the control of robots within the MuJoCo \cite{todorov2012mujoco} environment. 
In \cite{williams2017information}, a NN is used to learn the dynamics of a ground vehicle before applying it in a sampling-based MBRL framework. An overarching theme in these works is that while these learned models have been shown to be accurate in representing complex dynamics, they often require relatively large architectures, with a high number of hidden layers, neurons or models within the ensemble. Hence, it is challenging to use these methods within a conventional nonlinear MPC formulation \cite{borrelli2017predictive}, in which a constrained nonlinear optimization problem is solved at every time step during deployment.  

One possible solution is to utilize the KNODE-MPC framework proposed in \cite{chee2022knode}, where a neural ordinary differential equation (NODE) model is used to characterize the residual dynamics of a quadrotor system. The NODE model is combined with a first principles model to form a knowledge-based NODE (KNODE) model, which is then employed within an MPC framework. Because of its lightweight architecture and sample efficiency, the KNODE-MPC framework is more flexible and allows for practical extensions \cite{jiahao2022online, chee2022learning}. 

Another drawback of these dynamics learning methods is that the training of the model typically happens offline. As a result, the disturbances experienced by the robot during operation are not accounted for in the training process. Although there are recent studies that attempt to bridge this gap by introducing online and active learning methods, \textit{e.g.}, \cite{jiahao2022online, saviolo2022active}, they tend to be computationally heavy and require a relatively large amount of data for training. This potentially prohibits the robot from adapting to fast-changing disturbances within dynamic environments.

To improve the adaptability and control performance of the quadrotor system against uncertainties, there exist techniques that combine adaptive control with MPC. The authors in \cite{pereida2018} combine $\calL_1$ adaptive control with MPC to improve the closed-loop performance of a quadrotor system, albeit in a linear setting. In \cite{hanover2021performance}, the authors apply $\calL_1$ adaptive control on top of a nonlinear MPC scheme. Along the same vein, the authors in \cite{wu20221}, \cite{pravitra2020} and \cite{cheng2022} apply a similar $\calL_1$ adaptive module to geometric control, MBRL, and to a set of RL algorithms. It is important to note that in these works, the $\calL_1$ adaptive module is structured as an additive component where the $\calL_1$ control signal is directly added to a signal of an existing controller. Moreover, the $\calL_1$ control signal only accounts for  
matched uncertainties, 
a subset of all possible uncertainties. Matched uncertainties refer to uncertainties  that enter the system through the same channel as the control input. For the quadrotor system, matched uncertainties are the unmodeled forces along the body $z$ axis, in the same direction as the thrust generated by the motors, as well as unmodeled moments about all three axes. On the other hand, unmatched uncertainties are the unmodeled forces along the body $x$ and $y$ axes \cite{wu20221}. 

\textit{Contributions:} Our contributions in this work are four-fold. First, we propose a novel learning-enhanced MPC framework, $\calL_1$-KNODE-MPC, which compensates for both matched and unmatched uncertainties in a sample-efficient manner during deployment. Second, to the best of the authors' knowledge, this is the first work that combines both Neural ODEs and $\calL_1$ adaptive control methods within a nonlinear constrained MPC framework and demonstrates its application on quadrotor control. A schematic of the framework is depicted in Fig. \ref{fig:schematic}. Third, we provide an important insight on how both matched and unmatched uncertainties can be incorporated into the dynamics model within KNODE-MPC. We leverage this insight in the formulation of our second variant, $\calL_1$-KNODE-MPC-\textit{Int}. Lastly, through simulations and physical experiments, we demonstrate that $\calL_1$-KNODE-MPC not only allows the uncertainties to be estimated accurately, but also enables the quadrotor system to achieve significant performance improvements in terms of trajectory tracking.
\vspace{-0.1cm}
\section{PRELIMINARIES}
\vspace{-0.1cm}
We first describe the dynamics of a quadrotor system and a learning-based nonlinear MPC framework, KNODE-MPC.
\subsection{Quadrotor Dynamics}
We consider a quadrotor system with the following dynamics \cite{mellinger2011minimum},
\vspace{-0.15cm}
\begin{equation}\label{eq:eom}
\small \vspace{-0.15cm} \begin{split}
    \dot{\bfr} = \bfv,\;\;\,\quad\qquad\qquad
    m\dot{\bfv} &= -m\bfg + \bfR\mathbf{e}_3\, \eta,\\
    \bfJ\dot{\boldsymbol{\omega}} = \boldsymbol{\tau} - \boldsymbol{\omega} \times \bfJ \boldsymbol{\omega},\qquad
    \dot{\bfq} &= \frac{1}{2} \boldsymbol{\Omega} \bfq,
\end{split}
\end{equation}
where $\bfr,\,\bfv,\,\boldsymbol{\omega}\in \setR^3$ and $\bfq \in \setR^4$ are the position, velocity, angular rate and quaternions representing the dynamics and kinematics of the quadrotor. The parameters $\eta \in \mathbb{R}$ and $\boldsymbol{\tau} \in \mathbb{R}^3$ are the thrust and moments generated by the motors of the quadrotor. The vector $\mathbf{g}$ denotes gravity in the inertial reference frame and $\bfR \in \mathbb{R}^{3\times3}$ is the transformation matrix from the body to the inertial reference frame. The inertial frame follows the East-North-Up convention where the $z$ axis is pointing upwards. The unit vector $\mathbf{e}_i\in \mathbb{R}^{3}$ has value one in the $i^{th}$ element and zero elsewhere. The matrix $\boldsymbol{\Omega} \in \setR^{4\times 4}$ is defined as
\vspace{-0.3cm}
\begin{equation} 
\small 
    \boldsymbol{\Omega} := \begin{bmatrix}0 & -\omega_x & -\omega_y & -\omega_z\\\omega_x& 0 & \omega_z & -\omega_y\\\omega_y & -\omega_z & 0 & \omega_x\\
    \omega_z & \omega_y & -\omega_x & 0\end{bmatrix},
\end{equation}
where $\omega_x,\,\omega_y,\,\omega_z$ are components of the angular rate $\boldsymbol{\omega}$. The mass and inertia matrix of the quadrotor are denoted by $m$ and $\bfJ\in \setR^{3\times3}$ respectively. By defining the state and control inputs as $\bfx := [\bfr^{\top}\;\bfv^{\top}\;\bfq^{\top}\;\boldsymbol{\omega}^{\top}]^{\top}$ and $\bfu := [\eta\; \boldsymbol{\tau}^{\top}]^{\top}$, the equations of motion \eqref{eq:eom} can be written in a compact form, 
\vspace{-0.2cm}
\begin{equation} \label{eq:compact_eom}
    \dot{\bfx} = \mathbf{f}(\bfx,\bfu),
\vspace{-0.2cm}
\end{equation}
which we refer to as the nominal dynamics model. 
\vspace{-0.1cm}
\subsection{KNODE-MPC}
\vspace{-0.1cm}
For applications that do not require a high-fidelity model, the nominal dynamics model in \eqref{eq:compact_eom} may be of sufficient accuracy. However, when it is applied within a model-based control framework such as nonlinear MPC, it is challenging to ascertain if the nominal model is accurate enough, especially in the presence of unmodeled dynamics such as actuator dynamics and structural vibrations \cite{nonami2010autonomous}. On the other hand, when the quadrotor is deployed, we can collect data that provide information about the true system dynamics. The KNODE framework \cite{jiahao2021knowledge} leverages the knowledge of the nominal model, as well as the collected data, to construct an accurate dynamics representation, known as the KNODE model. This model is then applied within an MPC framework, collectively known as KNODE-MPC, to improve the closed-loop tracking performance of the quadrotor system \cite{chee2022knode, jiahao2022online}. Within KNODE-MPC, we consider the following constrained optimization problem, 
\vspace{-0.2cm}
\begin{subequations} \label{eq:ftocp}
\begin{align}
    \underset{\bfu}{\textnormal{minimize}}\quad\; & \sum_{i=0}^{N-1} ||\bfx_i- \bfx_{r,i}||^2_Q + ||\bfu_i||_R^2 \\
    &\quad + ||\bfx_N-\bfx_{r,N}||^2_P\\
    \text{subject to}\quad\; &\bfx_{i+1} = \mathbf{f}_{\boldsymbol{\theta},d}(\bfx_i, \bfu_i), \quad \forall\, i\in [0,\,N-1] \label{eq:dynamics}\\ 
    \; &\bfx_i \in \mathcal{X}, \quad \bfu_i \in \mathcal{U}, \quad \forall\, i\in [0,\,N-1]\\
    \; &\bfx_N \in \mathcal{X}_f,\quad \bfx_0 = \bfx(k),
\end{align}
\end{subequations}
where $N$ is the prediction horizon, $\mathcal{X},\,\mathcal{X}_f,\,\mathcal{U}$ are sets in which state and control input constraints are defined. The matrices $Q,\,P$ and $R$ are weighting matrices for the stage, terminal and control input costs respectively. For a vector $\bfs$ and matrix $\bfA$, $||\bfs||^2_{\bfA}$ denotes $\bfs^{\top}\bfA\bfs$. The vector $\bfx(k)$ is the state measurement obtained at each time step $k \in \setN$ and $\{\bfx_{r,0},\dotsc,\bfx_{r,N}\}$ is a sequence of reference states. At each time step $k$, we solve \eqref{eq:ftocp} and obtain a sequence of optimal control inputs, $\bfu^{\star} := \{\bfu_0^{\star},\dotsc,\bfu_{N-1}^{\star}\}$. The first vector in this sequence $\bfu_0^{\star}(\bfx(k))$ is then applied to the system as the control action. 

The model $\mathbf{f}_{\boldsymbol{\theta},d}(\bfx,\bfu)$ in \eqref{eq:dynamics} is a discrete-time version of the KNODE model, which, with a slight abuse of notation, is given as
\vspace{-0.4cm}
\begin{equation} \label{eq:knode_model}
\vspace{-0.2cm}
\dot{\bfx} = \mathbf{f}_{\boldsymbol{\theta}}(\bfx,\bfu) := \mathbf{f}(\bfx,\bfu) + \mathbf{d}_{\boldsymbol{\theta}}(\bfx,\bfu),
\end{equation}
where $\mathbf{f}(\bfx,\bfu)$ is the nominal model in \eqref{eq:compact_eom} and $\mathbf{d}_{\boldsymbol{\theta}}(\bfx,\bfu)$ is a neural network with parameters $\boldsymbol{\theta}$ representing the learned residual dynamics. Further details on the data collection process and training of the KNODE model can be found in \cite{chee2022knode}. With a suitable choice of the cost matrix $P$ and constraint set $\mathcal{X}_f$, sufficient conditions for asymptotic stability of the closed-loop system can be attained \cite{chee2022learning}.

While the neural network $\mathbf{d}_{\boldsymbol{\theta}}(\bfx,\bfu)$ is able to characterize the residual dynamics that are not accounted for by the nominal model, it is important to note that training of the KNODE model happens offline. This implies that the KNODE model is only able to account for uncertainties that manifest within the collected data, and not those in subsequent deployments. This motivates us to incorporate the components of $\calL_1$ adaptive control to account for uncertainties that are not compensated by the KNODE model.

\vspace{-0.2cm}
\section{INCORPORATING $\calL_1$ ADAPTATION}
\vspace{-0.1cm}
\subsection{$\calL_1$ Adaptive Control}
\vspace{-0.1cm} For the control of a quadrotor system, existing state-of-the-art frameworks use $\calL_1$ adaptive control as a separate, additive module to a baseline control scheme, \textit{e.g.}, \cite{hanover2021performance, wu20221}. 
The $\calL_1$ adaptive control structure consists of three sub-modules; a state predictor, an adaptation law and a low-pass filter. The state predictor estimates the mismatch between a model and the true dynamics in the form of matched and unmatched uncertainties. By defining a partial state vector $\bfz:=[\bfv^{\top}\,\boldsymbol{\omega}^{\top}]^{\top}$ and considering the residual dynamics in the KNODE model, $\bfd_{\boldsymbol{\theta}}(\bfx, \bfu)$, the state predictor is given as
\vspace{-0.3cm}
\begin{equation} \label{eq:state_predictor}
\small \vspace{-0.3cm} \begin{split}
    \dot{\hat{\bfz}} &= \begin{bmatrix} -\mathbf{g} \\ -\bfJ^{-1}(\boldsymbol{\omega} \times \bfJ\boldsymbol{\omega}) \end{bmatrix} + \bfd_{\boldsymbol{\theta},\bfz}(\bfx,\mathbf{\bar{u}}) + \bfG_1 \mathbf{\bar{u}}
    +\bfG \hat{\boldsymbol{\sigma}} + \mathbf{A}(\hat{\bfz}-\bfz),
\end{split}
\end{equation}
where $\mathbf{\bar{u}}:=\bfu_b +\bfu_{\calL_1}$, $\bfu_b$ and $\bfu_{\calL_1}$ are the control inputs from the baseline control scheme and $\calL_1$ control law respectively.  The vector $\hat{\boldsymbol{\sigma}} := [\hat{\boldsymbol{\sigma}}_m^{\top}\; \hat{\boldsymbol{\sigma}}_{um}^{\top}]^{\top} \in \mathbb{R}^6$ consists of the estimates of the matched uncertainties $\hat{\boldsymbol{\sigma}}_m \in \mathbb{R}^4$ and the unmatched uncertainties $\hat{\boldsymbol{\sigma}}_{um} \in \mathbb{R}^2$. The state-dependent matrix $\bfG := [\bfG_1\,\bfG_2] \in \mathbb{R}^{6 \times 6}$ describes the mapping between the uncertainties and the predicted state, with the matrices $\bfG_1 \in \mathbb{R}^{6 \times 4}$ and $\bfG_2 \in \mathbb{R}^{6 \times 2}$ defined as
\vspace{-0.1cm}
\begin{equation}
\vspace{-0.1cm}
\bfG_1 := \Big[ \begin{smallmatrix} (1/m) \bfR\mathbf{e}_3 & \mathbf{0}_{3\times3}\\ \mathbf{0}_{3\times1} & \bfJ^{-1}\end{smallmatrix} \Bigr],\; \bfG_2 := \Bigl[\begin{smallmatrix} (1/m) \bfR\mathbf{e}_1 & (1/m) \bfR\mathbf{e}_2\\ \mathbf{0}_{3\times1} & \mathbf{0}_{3\times1}\\ \end{smallmatrix}\Bigr].
\end{equation}
The matrix $\mathbf{A}$ is a pre-specified diagonal Hurwitz matrix. This formulation is similar to the one in \cite{wu20221}, except for the vector $\bfd_{\boldsymbol{\theta},\bfz}(\bfx,\mathbf{\bar{u}}) \in \setR^{6}$. This vector is a subset of the residual dynamics in the KNODE model \eqref{eq:knode_model} that corresponds to the partial state $\bfz$. The inclusion of $\bfd_{\boldsymbol{\theta},\bfz}(\bfx,\mathbf{\bar{u}})$ ensures that only uncertainties that have not been considered within the KNODE model are estimated by the adaptation law.
Next, the adaptation law is formulated in a piecewise constant manner \cite{wu20221} such that for $t \in [kT,\,(k+1)T]$,
\vspace{-0.15cm}
\begin{equation} \label{eq:adapt_law}
\vspace{-0.15cm}
\begin{split}
    \hat{\boldsymbol{\sigma}}(t) &:= \hat{\boldsymbol{\sigma}}(kT)
    := \bfG^{-1} \left(e^{\bfA T} - \mathbf{I}\right)^{-1} \bfA e^{\bfA T} (\bfz(k)-\hat{\bfz}(k)),
\end{split}
\end{equation}
where $k \in \setN$, $T$ is the sampling time step and $\mathbf{I}$ is the identity matrix with appropriate dimensions. Finally, the $\calL_1$ control law is constructed by accounting for matched uncertainties that are within the bandwidth of the specified low-pass filter \cite{wu20221}. In the Laplace domain, the control signal $\bfu_{\calL_1}$ is written as
\vspace{-0.25cm}
\begin{equation} \label{eq:l1_control_law}
\vspace{-0.2cm}
    \bfu_{\calL_1}(s) = -\bfF(s)\hat{\boldsymbol{\sigma}}_m(s),
\end{equation}
where $\bfF(s)$ is the transfer function of the low-pass filter. 

We highlight that the control signal in \eqref{eq:l1_control_law} only accounts for matched uncertainties. In Section \ref{sec:l1_knode_mpc}, we present an important insight and show that it is possible and beneficial to incorporate \emph{both} the matched and unmatched uncertainties into the KNODE-MPC framework.

\vspace{-0.2cm}
\subsection{$\calL_1$-KNODE-MPC} \label{sec:l1_knode_mpc}
\vspace{-0.15cm}
In our proposed 
framework, we present two variants. In the first variant, $\calL_1$-KNODE-MPC-\textit{Direct}, we consider KNODE-MPC as the baseline control scheme and incorporate $\calL_1$ adaptive control \emph{directly} as an add-on module. Specifically, we add the control signal obtained from KNODE-MPC, \textit{i.e.}, $\bfu^{\star}_0(\bfx(k))$, with the discretized adaptive control signal $\bfu_{\calL_1}(k)$ from \eqref{eq:l1_control_law}. 
In this variant, the combination of $\calL_1$ adaptive control and KNODE-MPC occurs at the level of the control inputs. This coupling can be seen as an intuitive, albeit weaker integration of the two components, and does not account for unmatched uncertainties. Nonetheless, it shows the feasibility of such a combination.
This is depicted in the top panel of Fig. \ref{fig:schematic}.
In our second variant, $\calL_1$-KNODE-MPC-\textit{Int}, shown in the bottom panel of Fig. \ref{fig:schematic}, we \emph{integrate} the uncertainties estimated from the adaptation law into the KNODE model \eqref{eq:knode_model}. We observe from \eqref{eq:state_predictor} and \eqref{eq:adapt_law} that the product of the matrix $\bfG$ and 
$\boldsymbol{\hat{\sigma}}$ can be interpreted as part of the system dynamics, and in particular, part of the residual translational and rotational accelerations. This observation allows us to formulate the piecewise-constant uncertainties as residual accelerations acting on the quadrotor such that for $t \in [kT,\,(k+1)T]$,
\vspace{-0.15cm}
\begin{equation} \label{eq:l1_uncertainty}
\vspace{-0.15cm}
\begin{split}
    \begin{bmatrix} \bff_{\boldsymbol{\sigma}}(t)\\ \bfm_{\boldsymbol{\sigma}}(t) \end{bmatrix} &:= \begin{bmatrix} \bff_{\boldsymbol{\sigma}}(kT)\\ \bfm_{\boldsymbol{\sigma}}(kT) \end{bmatrix}
    := \bfG \hat{\boldsymbol{\sigma}}(kT),\\
\end{split}
\end{equation}
where $\bff_{\boldsymbol{\sigma}}(t),\,\bfm_{\boldsymbol{\sigma}}(t) \in \setR^3$ are the translational and rotational uncertainties. We then combine these uncertainties with the KNODE model \eqref{eq:knode_model} to form the following $\calL_1$-KNODE model,
\vspace{-0.4cm}
\begin{equation} \label{eq:l1_knode_model}
\vspace{-0.1cm}
\begin{split}
\dot{\bfx} := \bff_{\boldsymbol{\theta}}(\bfx,\bfu) + \hat{\bfd}(\bfx,\bfu) := \bff_{\boldsymbol{\theta},\calL_1}(\bfx,\bfu),
\end{split}
\end{equation}
where $\hat{\bfd}(\bfx,\bfu) := [\mathbf{0}_{1\times 3}\; \bff_{\boldsymbol{\sigma}}(t)^{\top}\; \mathbf{0}_{1\times 4}\; \bfm_{\boldsymbol{\sigma}}(t)^{\top}]^{\top}$. With this combination, we augment the data-driven KNODE model to include the estimated uncertainties from the adaptation law. This augmented model is then used as the dynamics model \eqref{eq:dynamics} within KNODE-MPC. 
We highlight that these uncertainties are estimated using the current state, and are not computed in conjunction with the predicted states in \eqref{eq:ftocp}. They are integrated into the dynamics constraints \eqref{eq:dynamics} after the estimation procedure, and act as parameters to \eqref{eq:ftocp}. This allows the MPC scheme to account for these estimated uncertainties during the optimization procedure. We summarize the two proposed variants of $\calL_1$-KNODE-MPC in Algorithms \ref{alg:algo1} and \ref{alg:algo2}. 

In $\calL_1$-KNODE-MPC-\textit{Int}, there is no segregation between matched and unmatched uncertainties. Both types of uncertainties are considered, which is in stark contrast to existing $\calL_1$ adaptive control frameworks. This allows the MPC scheme to generate control inputs that account for both sources of uncertainties. For instance, if the quadrotor experiences an uncertain force in the body $x$ axis, the inclusion of an estimate of this force into the dynamics model, across the prediction horizon, allows the MPC scheme to orientate the quadrotor with a non-zero pitch angle and adjust the motor thrust to counteract this uncertain force, while compensating for gravitational forces. Hence, even though the quadrotor is not able to actuate in the body $x$ axis directly, it is still able to account for disturbances in that direction. 

Another advantage of this second variant is that it resolves the inconsistency between the dynamics model used in MPC and that of the $\calL_1$ state predictor, as pointed out in \cite{saviolo2022active}. Both dynamics models consider the residual dynamics estimated by the NODE model $\bfd_{\boldsymbol{\theta}}(\bfx,\bfu)$, as well as the dynamics estimated by the adaptive module, $\hat{\bfd}(\bfx,\bfu)$.

\begin{figure}
\vspace{0.05cm}
\removelatexerror
\begin{algorithm}[H] \label{alg:algo1}
\DontPrintSemicolon
  \KwInput{KNODE model $\bff_{\boldsymbol{\theta}}$, state measurements $\bfx(k)$,}
  \KwOutput{Control action $\mathbf{\bar{u}}(k)$} 
  Initialize: $\bar{\bfu} \leftarrow \bf0$, $\hat{\bfz} \leftarrow \bfz$\;
  \textit{At each time step $k \in \setN$:}\;
  \Indp Extract $\bfd_{\boldsymbol{\theta},\bfz}$ from $\bff_{\boldsymbol{\theta}}$ in \eqref{eq:knode_model}\;
  Discretize and propagate \eqref{eq:state_predictor} to get $\hat{\bfz}(k)$\;
  Compute uncertainties $\hat{\boldsymbol{\sigma}}(k)$ using \eqref{eq:adapt_law}\;
  Compute $\bfu_{\calL_1}(k)$ with \eqref{eq:l1_control_law} or \eqref{eq:discrete_lpf}\;
  Discretize KNODE model $\bff_{\boldsymbol{\theta}}$ in \eqref{eq:knode_model} and set it as $\bff_{\boldsymbol{\theta},d}$ in \eqref{eq:dynamics}\;
  Apply $\bfx(k)$ to solve \eqref{eq:ftocp} and get $\bfu^{\star}_0(\bfx(k))$\;
  Compute $\mathbf{\bar{u}}(k)\leftarrow\bfu^{\star}_0(\bfx(k))+\bfu_{\calL_1}(k)$
\caption{$\calL_1$-KNODE-MPC-\textit{Direct}}
\end{algorithm}
\end{figure}

\begin{figure}
\vspace{-0.75cm}
\removelatexerror
\begin{algorithm}[H] \label{alg:algo2}
\DontPrintSemicolon
  \KwInput{KNODE model $\bff_{\boldsymbol{\theta}}$, state measurements $\bfx(k)$,}
  \KwOutput{Control action $\bfu^{\star}_0(\bfx(k))$} 
  Initialize: $\bfu_0^{\star} \leftarrow \bf0$, $\hat{\bfz} \leftarrow \bfz$\;
  \textit{At each time step $k \in \setN$:}\;
  \Indp Extract $\bfd_{\boldsymbol{\theta},\bfz}$ from $\bff_{\boldsymbol{\theta},\calL_1}$ in \eqref{eq:l1_knode_model}\;
  Discretize and propagate \eqref{eq:state_predictor} to get $\hat{\bfz}(k)$\;
  Compute uncertainties $\hat{\boldsymbol{\sigma}}(k)$ using \eqref{eq:adapt_law}\;
  Compute $\hat{\bfd}(\bfx,\bfu)$ with \eqref{eq:l1_uncertainty} and \eqref{eq:l1_knode_model}\;
  Discretize $\calL_1$-KNODE model $\bff_{\boldsymbol{\theta},\calL_1}$ in \eqref{eq:l1_knode_model} and set it as $\bff_{\boldsymbol{\theta},d}$ in \eqref{eq:dynamics}\;
  Apply $\bfx(k)$ to solve \eqref{eq:ftocp} and get $\bfu^{\star}_0(\bfx(k))$\;
\caption{$\calL_1$-KNODE-MPC-\textit{Int}}
\end{algorithm}
\end{figure}
\setlength{\textfloatsep}{-0.05cm}

From a broader perspective, the $\calL_1$-KNODE-MPC framework utilizes the expressiveness of the deep learning component, \textit{i.e.}, the NODE model, to primarily account for systematic uncertainties, whose effects are likely to be present in the training data. The adaptive module in the framework is leveraged upon to account for time-varying external uncertainties experienced by the robot during deployment in an online fashion.
\vspace{-0.25cm}
\subsection{Implementation Details}
\vspace{-0.1cm}We use the Optistack class within the CasADi package \cite{Andersson2019} to implement the optimization problem in \eqref{eq:ftocp}. A solver based on the interior-point method, IPOPT \cite{wachter2006implementation}, is used to solve \eqref{eq:ftocp}.
Discretization of the KNODE model \eqref{eq:knode_model}, $\calL_1$-KNODE model \eqref{eq:l1_knode_model} is performed using the explicit fourth-order Runge-Kutta (RK) method, while the state predictor in \eqref{eq:state_predictor} is discretized with the fifth-order RK method \cite{shampine1986some}. The filter \eqref{eq:l1_control_law} is implemented as
\vspace{-0.15cm}
\begin{equation} \label{eq:discrete_lpf}
    \vspace{-0.15cm}
    \bfu_{\calL_1}(k) = \big(\bfu_{\calL_1}(k-1) + \hat{\boldsymbol{\sigma}}_m(kT)\big)e^{-\gamma T} - \hat{\boldsymbol{\sigma}}_m(kT),
\end{equation}
for all $k \in \setN$ and $\gamma$ is the cut-off frequency. Training of the KNODE model is done in PyTorch using \textit{torchdiffeq} \cite{chen2018neuralode}.
\vspace{-0.1cm}
\section{EXPERIMENTS AND RESULTS} \label{sec:results}
\vspace{-0.13cm} In our experiments and evaluation procedure, we set out to answer the following questions about our proposed framework: (i) How accurate are the uncertainties estimated by the $\calL_1$ adaptive module? (ii) How much performance improvement do the variants of the $\calL_1$-KNODE-MPC provide, in the presence of either matched, unmatched or both types of uncertainties? (iii) Can such performance improvements observed in simulations be carried forward and validated in physical experiments? To address the first question, we apply time-varying uncertainties to the quadrotor in simulation and compare the uncertainty estimates against the ground truth. For the other two questions, we conduct tests in both simulations and physical experiments and compare the proposed variants against four benchmarks. 

The first benchmark is a nonlinear MPC framework. It uses the nominal model in \eqref{eq:compact_eom} and does not account for any uncertainties.
The second benchmark is KNODE-MPC \cite{chee2022knode}. Comparison with this benchmark allows us to quantify the improvement brought forth by $\calL_1$ adaptation. As a third benchmark, we consider a non-learning
MPC framework that directly includes the $\calL_1$ adaptive module, which we refer to as $\calL_1$-MPC. A similar framework is presented in \cite{hanover2021performance}. This framework combines the $\calL_1$ adaptation at the level of the control input signals, similar to $\calL_1$-KNODE-MPC-\textit{Direct}, but does not consider any learning or data-driven components. The last benchmark that we consider is KNODE-MPC-Online \cite{jiahao2022online}. This framework uses online learning to update the dynamics model using data collected during robot deployment.  
\vspace{-0.25cm}
\subsection{Simulations} \label{sec:simulations}
\vspace{-0.15cm}
We first construct a quadrotor model using the equations given in \eqref{eq:compact_eom}. An explicit 5$^{\text{th}}$ order RK method (RK45) with a time step of 2 milliseconds is used for numerical integration to simulate dynamic responses for the quadrotor system. It is assumed that the model predictive controller and the $\calL_1$ adaptation module have access to the states of the quadrotor. The control commands are generated using the benchmarks and the proposed control algorithms. These act as inputs to the quadrotor model. The KNODE model is trained using data collected from flying on a circular trajectory of radius 3m and speed 1m/s. The NN in the KNODE model has one hidden layer with 32 neurons and uses the hyperbolic tangent activation function. It is important to note that the NN is only required to characterize the residual dynamics and not the full dynamics, and this enables the lightweight structure. Being lightweight is beneficial because it not only eases the training process in terms of sample efficiency, but also allows the optimization problem \eqref{eq:ftocp} to be solved efficiently. To test the efficacy of the KNODE model, we consider a nominal uncertainty in the form of a mass difference. The quadrotor has a true mass of 0.04kg, while the mass within the nominal model is set to be 0.03kg. The KNODE model is expected to account for this mass difference after training, but not other disturbances that occur during subsequent deployments. Additionally, we include time and state-dependent uncertainties during flight under the following test cases:
\vspace{-0.05cm}
\begin{itemize}
    \item Case 1: Roll moment disturbance 
    \vspace{-0.15cm}
    \begin{equation} \label{eq:roll_dist}
    \small d_{M_x}(t):=5\times10^{-4}\sin(0.75t) + 6\times10^{-4}, \end{equation}
    \vspace{-0.6cm}
    \item Case 2: Roll moment and side force disturbances
    \vspace{-0.15cm}
    \begin{equation} \label{eq:side_force_dist}
    \small
    d_{M_x}(t),\,
    d_{F_y}(t):=0.025(\sin(t) + 0.5\cos(1.5t) + 0.1t),
    \end{equation}
    \vspace{-0.6cm}
    \item Case 3: Case 2 with state-dependent disturbances
    \vspace{-0.1cm}
    \begin{equation} \label{eq:3d_force_dist}
    \small d_{M_x}(t), d_{F_y}(t),\,
    d_{F}(\bfx):= -\left({\bfC_d}/{m}\right) \,\text{sign}(\bfv_b)\, \bfv_b^2,
    \end{equation}
    \vspace{-0.45cm}
\end{itemize}
\vspace{-0.2cm}
where $\bfv_b \in \mathbb{R}^3$ are the velocities in the body reference frame, $\bfC_d \in \mathbb{R}^3$ are linear coefficients, with products computed element-wise. Case 1 considers only matched uncertainties, while Cases 2 and 3 include both matched and unmatched uncertainties. For the $\calL_1$ adaptive module, we set $\bfA := -\mathbf{I}$ and the low-pass filter to be the identity function in our simulations. 

\begin{figure}
    \centering
    {\vspace{0.2cm}}{\includegraphics[scale=0.26, trim = -0.1cm 1.5cm 0cm 0cm]{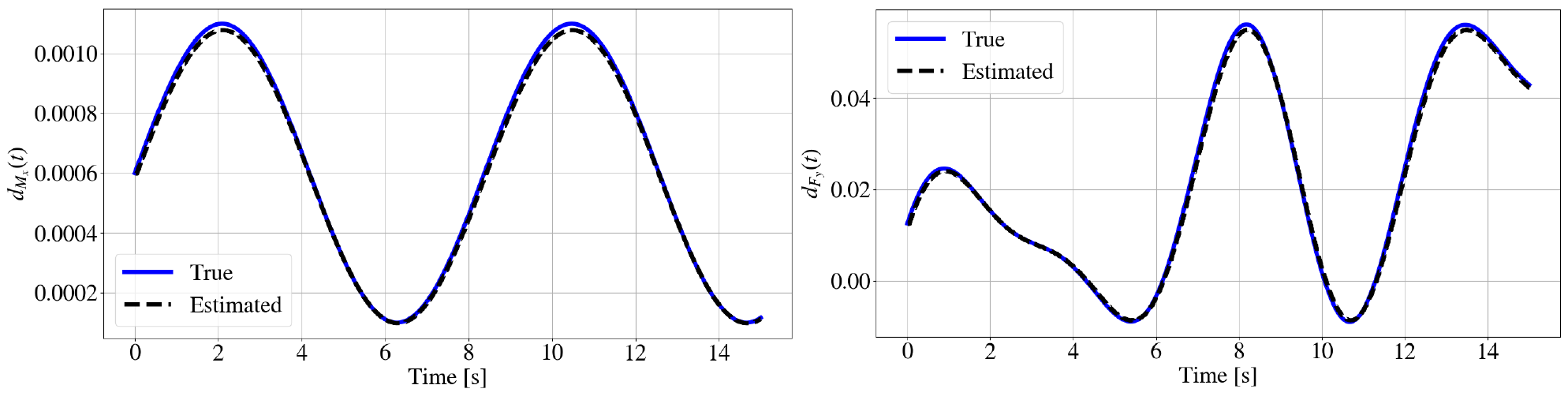}}
    \caption{\textbf{Uncertainty estimates:} Time histories of the uncertainties estimated by the adaptive module, against the true uncertainties. Left panel: Roll moment disturbance $d_{M_x}(t)$. Right panel: Side force disturbance $d_{F_y}(t)$.}
    \label{fig:uncertainty_est}
\end{figure}

\begin{figure}
    \centering
    {\includegraphics[scale=0.48, trim = 0.5cm 0.8cm 0cm -0.4cm]{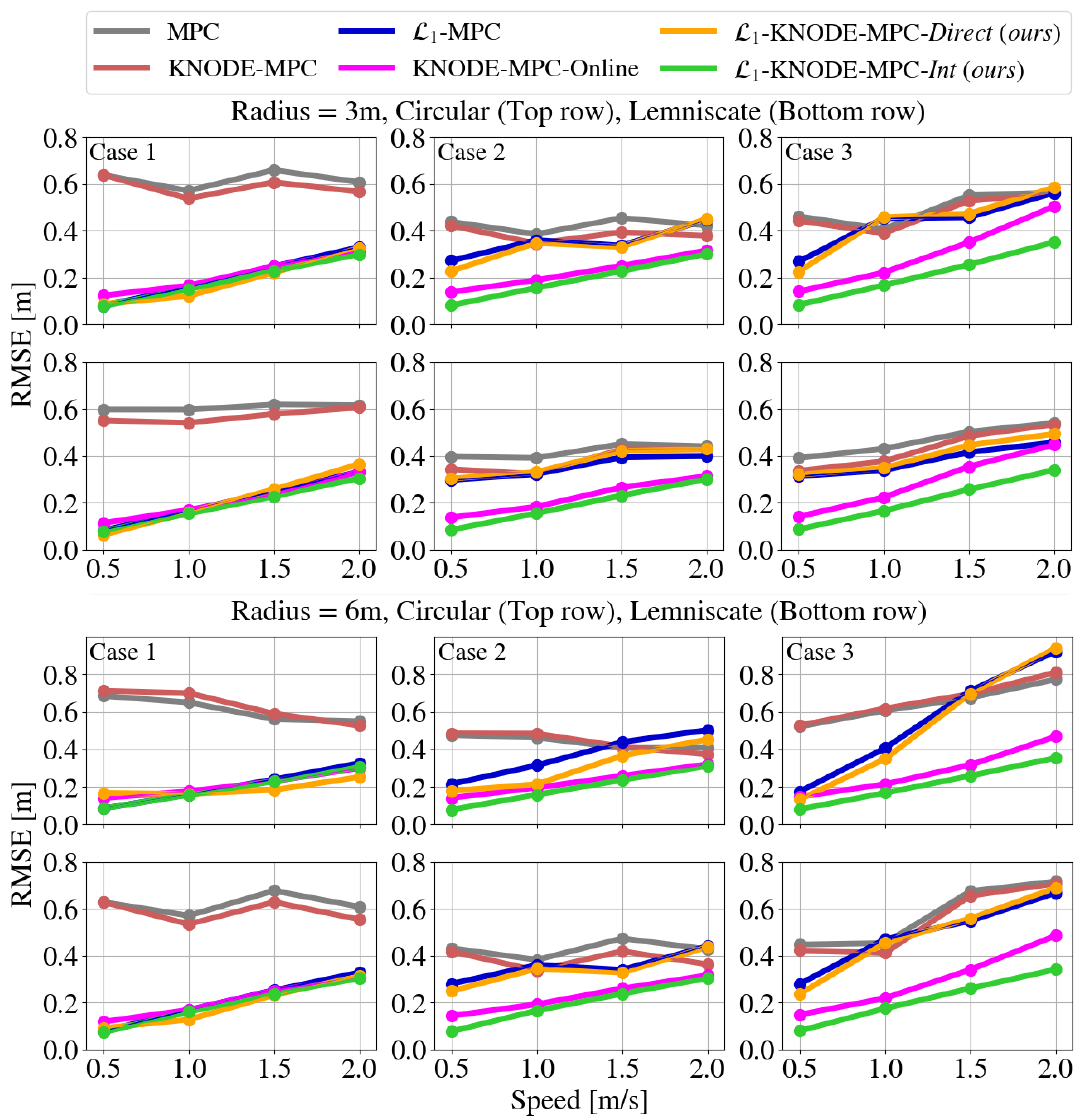}}
    \caption{\textbf{Simulation results}: Root mean squared errors (RMSEs) between the true and reference position vectors under the 3 test cases described in Section \ref{sec:simulations}. Simulations are conducted for both circular and lemniscate trajectories, across different radii and speeds.\vspace{-0.15cm}}
    \label{fig:sim_results}
\end{figure}

To ascertain the accuracy of the uncertainty estimates,
we apply the 
disturbances in Case 2 to the quadrotor separately, and compare 
the true and estimated uncertainties. As depicted in Fig. \ref{fig:uncertainty_est}, the adaptive module is able to estimate the disturbances accurately.
\setlength{\textfloatsep}{0cm}
Next, closed-loop simulations are conducted for each of the benchmarks and the proposed $\calL_1$-KNODE-MPC variants. For each of the test cases, we simulate the quadrotor with circular and lemniscate flight profiles of radii 3m and 6m, and at reference speeds ranging from 0.5 to 2 m/s. Each simulation has a flight duration of 15 seconds. To evaluate the trajectory tracking performance of the proposed framework, we compute the root mean squared errors (RMSE) between the true and reference position vectors. For the KNODE-MPC-Online approach, the RMSEs are obtained by taking the average across 10 runs. Results are shown in Fig. \ref{fig:sim_results}. In Case 1 where there are only matched uncertainties, the first proposed variant, $\calL_1$-KNODE-MPC-\textit{Direct} outperforms MPC, KNODE-MPC and $\calL_1$-MPC on average. In Cases 2 and 3, due to the presence of unmatched uncertainties, there is a noticeable degradation in performance for $\calL_1$-MPC and $\calL_1$-KNODE-MPC-\textit{Direct}. On the other hand, since $\calL_1$-KNODE-MPC-\textit{Int} accounts for \emph{both} matched and unmatched uncertainties, we observe consistently lower RMSEs across all test cases, at different speeds and radii, and for circular and lemniscate trajectories. This demonstrates the efficacy of the tighter integration between the $\calL_1$ adaptive module and KNODE-MPC within $\calL_1$-KNODE-MPC-\textit{Int}. Overall, $\calL_1$-KNODE-MPC-\textit{Int} achieves the best performance, with an average RMSE of 0.2m across all runs, outperforming the benchmarks by at least 18.1\%. We also observe that $\calL_1$-KNODE-MPC-\textit{Int} incurs a lower computational cost than KNODE-MPC-Online. The latter involves training a NN online, which is more computationally expensive than the computations required in the $\calL_1$ adaptive module.

\vspace{-0.15cm}
\subsection{Physical Experiments} \label{sec:expts}
\begin{figure}
    \centering
    {\includegraphics[scale=0.2, trim = 0cm 1.8cm 0cm -0.8cm]{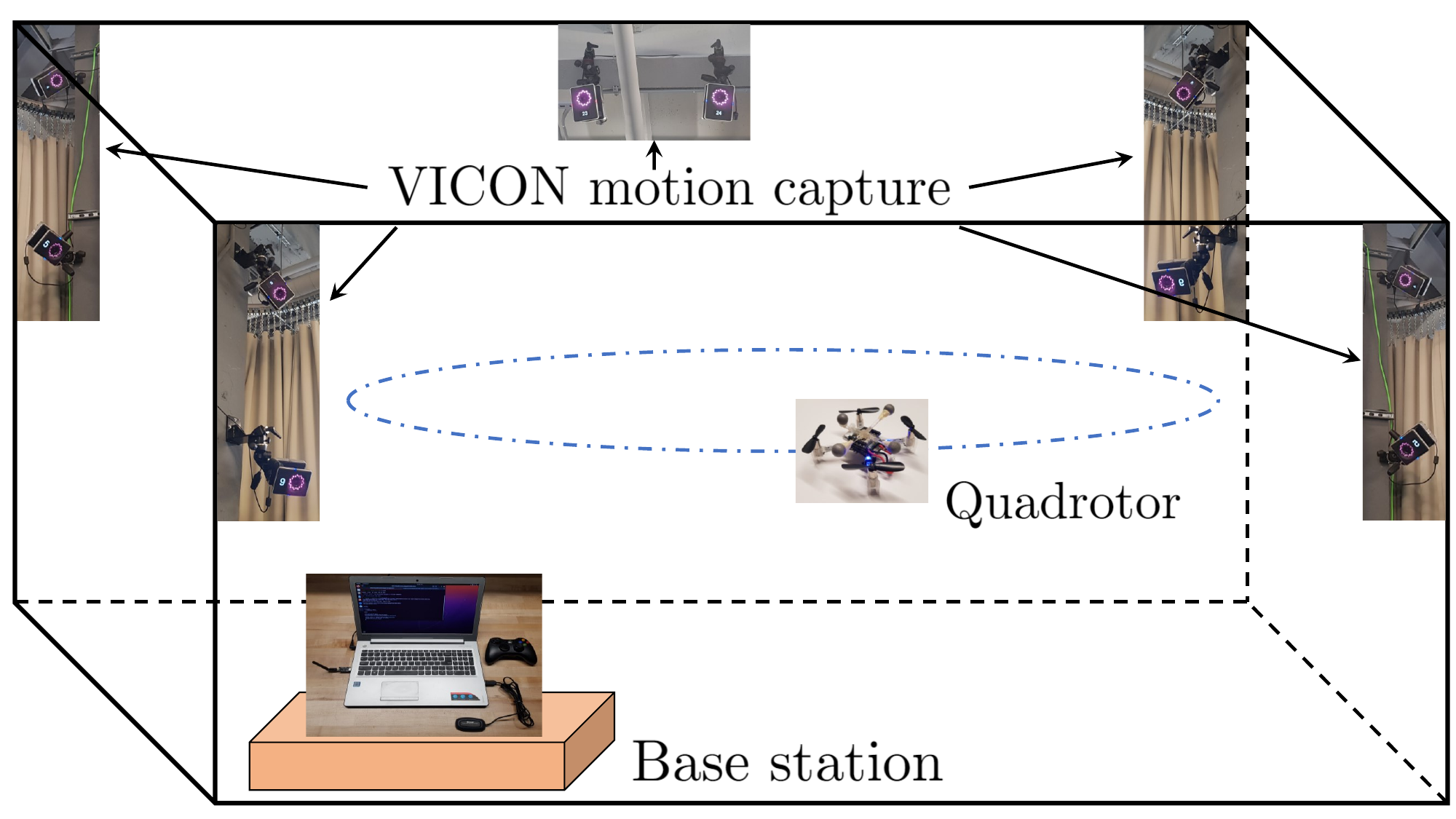}}
    \caption{\textbf{Experimental setup}: The open-source Crazyflie 2.1 quadrotor \cite{Bitcraze} is used as the experimental platform. A laptop running on Intel i7 CPU acts as the base station and communication with the Crazyflie is established via Crazyradio PA at an average rate of 300 Hz. A Vicon motion capture system is deployed to obtain pose information and communicates with the base station at an approximate rate of 100Hz. The CrazyROS wrapper \cite{crazyflieROS} is used as part of the software architecture. \vspace{-0.15cm}}
    \label{fig:expt_setup}
\end{figure}

\vspace{-0.1cm}
The setup for our physical experiments is depicted in Fig. \ref{fig:expt_setup}. The Crazyflie quadrotor has a size of 9 cm$^2$ and weighs approximately 34g. Linear velocities are estimated from the Vicon measurements through a combination of high and low pass filters, while accelerations and angular velocities are measured from the sensors onboard the quadrotor.
\setlength{\textfloatsep}{0.12cm}
For the training of the KNODE model, data is collected by flying the Crazyflie along a circular trajectory of radius 1m at a speed of 0.4m/s, for a duration of 25 seconds. The NODE model has 1 hidden layer of 4 neurons and uses the hyperbolic tangent activation function. The control frameworks run on the base station. These frameworks generate three-dimensional acceleration commands and run in tandem with a geometric controller \cite{mellinger2011minimum}, as well as attitude and thrust controllers within the Crazyflie firmware. The $\calL_1$ adaptive module is formulated by considering the velocities as states, \textit{i.e.}, $\bfz:=\bfv$. We set $\bfA := -2\mathbf{I}$ and the bandwidths of the low-pass filters to be 0.125, 0.125 and 0.75 rad/s for the three axes respectively. These filters are implemented for both proposed schemes in the physical experiments. Based on the experimental results, the estimated uncertainties improve closed-loop performance, even with the addition of the low-pass filter. To verify the adaptiveness of the framework, we attach a slung payload of mass 2g during flight at approximately 18.5 seconds after takeoff. We highlight that the addition of this slung payload manifests as both matched and unmatched uncertainties for the quadrotor system. The slung payload is not attached to the quadrotor rigidly and is allowed to move in the $x$ and $y$ axes. This effect is compounded in the second set of experiments where five pieces of payload are attached to one another at individual contact points. 
To get a better understanding of the physical experiments, we refer the reader to the accompanying video. 

\begin{figure}
    \centering
    {\includegraphics[scale=0.24, trim = 0.5cm 1.5cm 0cm -0.8cm]{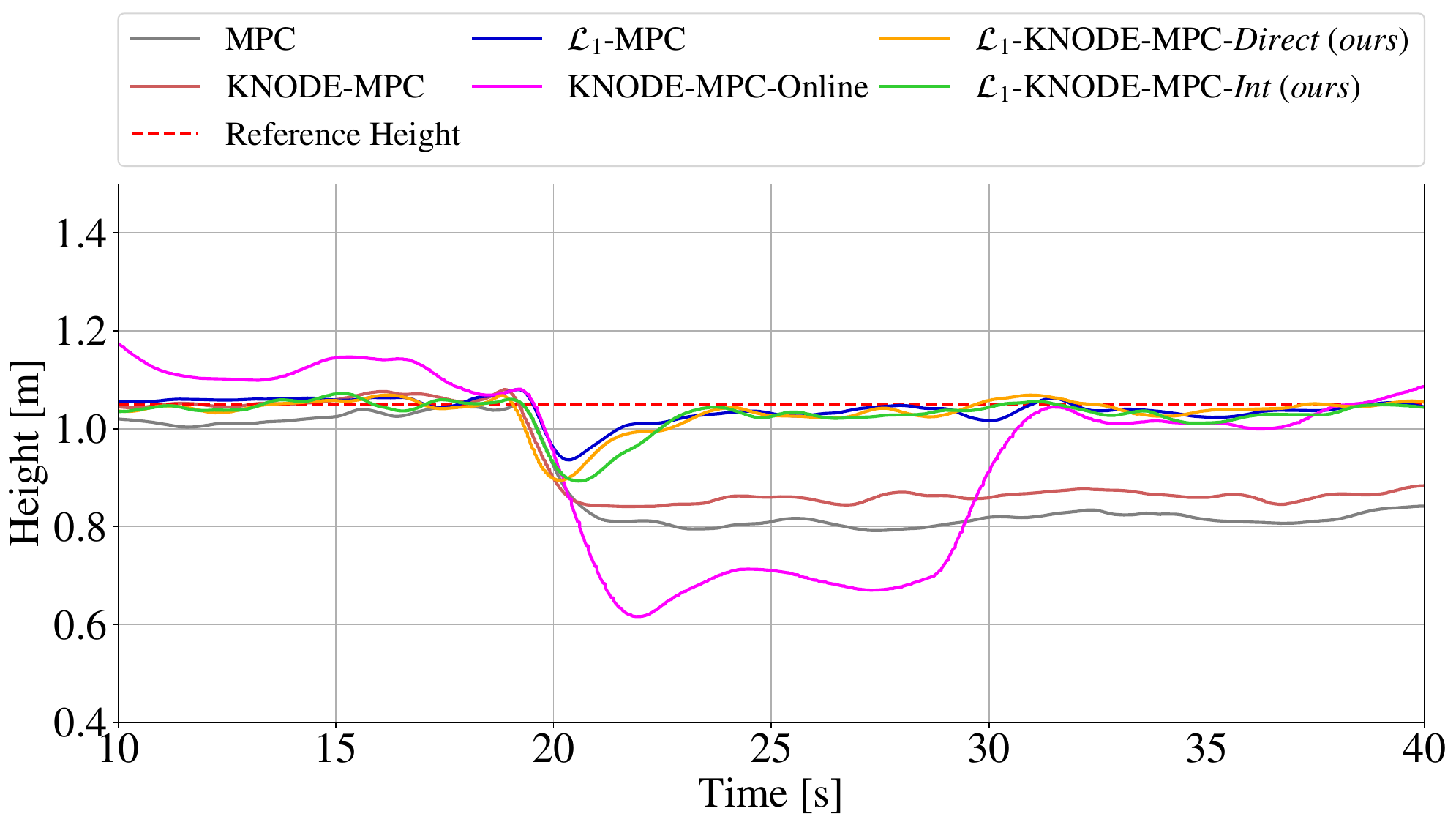}}
    \caption{\textbf{Height-time plots (i):} Time histories of the quadrotor height for the benchmarks and proposed variants during flight. A slung payload is attached to the quadrotor system at approximately t=18.5s.}
    \label{fig:phy_expt_height}
\end{figure}

\vspace{-0.05cm}
For each of the proposed methods and benchmarks, we conduct 10 runs along circular trajectories of radius 1m and speed of 0.4m/s to evaluate closed-loop trajectory tracking performance (a total of 60 runs). First, we investigate the effect of adaptation by plotting the time histories of the quadrotor height under the various control methods. As shown in Fig. \ref{fig:phy_expt_height}, the quadrotor experiences a drop in height when the slung payload is attached at t=18.5s. For KNODE-MPC, since the effects of this payload are not present in the training data, it does not account for those effects and the height does not recover to the original height. For KNODE-MPC-Online, because the data collection and training procedure takes time, it took approximately 11 seconds for the height to recover. For the methods with $\calL_1$ adaptation, the height recovers within approximately 4 seconds. 

% ...

\begin{figure}
    \centering
    {\includegraphics[scale=0.26, trim = 1cm 1cm 0cm 1.5cm]{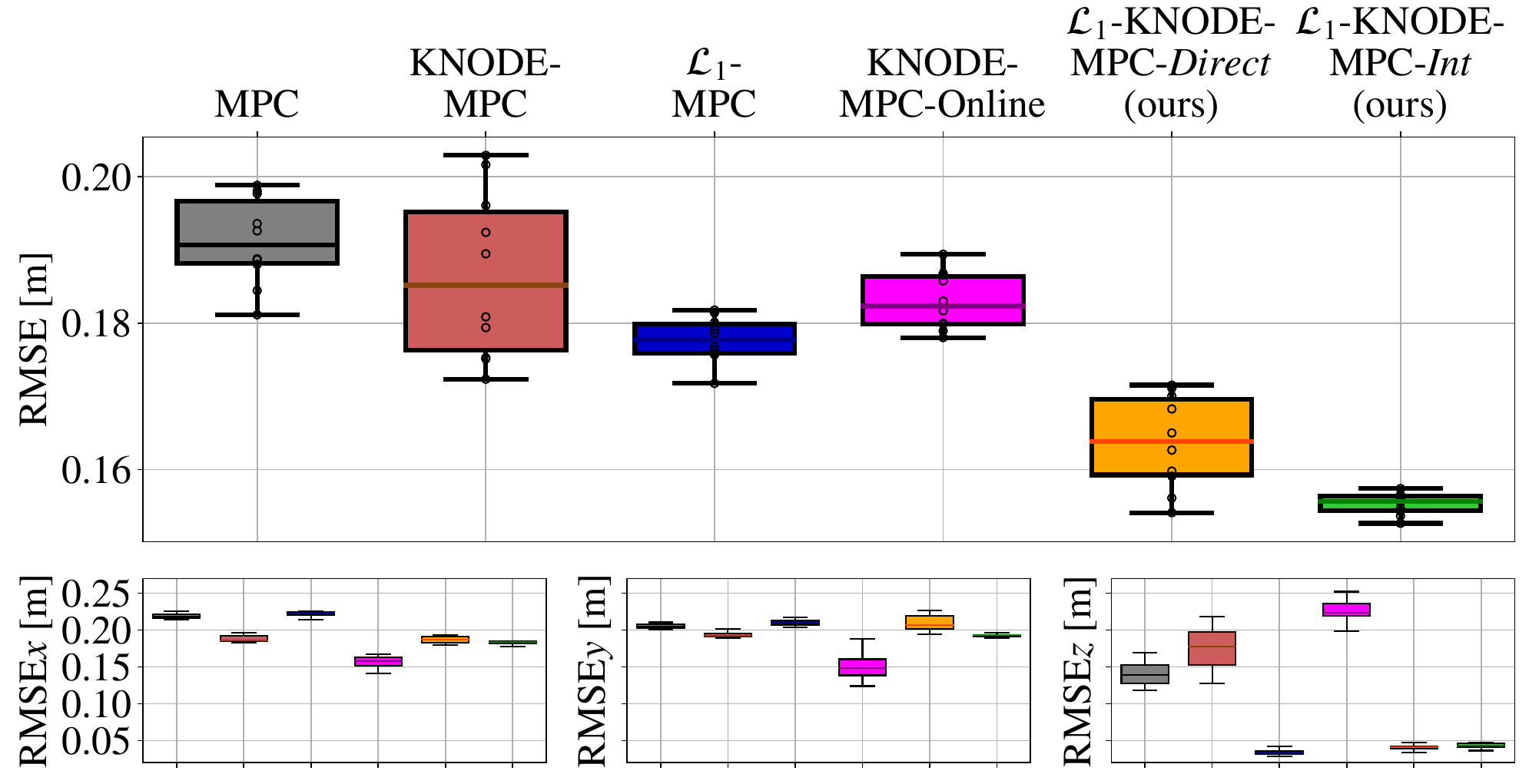}}
    \caption{\textbf{Experimental results:} Statistics of the RMSEs for the various MPC frameworks, with a slung payload attached during flight. For each of the box plots, the center line denotes the median and the height of the boxes denote the inter-quantile range. The RMSEs for the individual runs are marked with black circles.}
    \label{fig:phy_expt_plots}
\end{figure}

\vspace{-0.05cm}
Next, to evaluate the overall performance, we compute the RMSEs between the measured and reference position vectors in all 3 axes. Fig. \ref{fig:phy_expt_plots} depicts the statistics of the RMSEs. The proposed variants, $\calL_1$-KNODE-MPC-\textit{Direct} and \textit{Int}, achieve significant improvements over the benchmarks. Specifically, the median RMSEs for the two variants are 14.1\% and 18.4\% lower than that of MPC. Furthermore, the variance for $\calL_1$-KNODE-MPC-\textit{Int} is 76.9\% smaller than that of nominal MPC, in terms of inter-quantile ranges. This implies that $\calL_1$-KNODE-MPC-\textit{Int} provides more accurate and consistent performance in terms of trajectory tracking. More generally, we observe a reduction in variance when adaptation is incorporated into the control scheme. This implies that the quadrotor is able to achieve more consistent trajectory tracking with adaptation. The performance improvement obtained from the second variant over the first variant validates the tighter integration between the adaptive module and KNODE-MPC, and in particular, with the formulation of the $\calL_1$-KNODE model. 

\begin{figure}
    \centering
    {\includegraphics[scale=0.24, trim = -0.2cm 1.6cm 0cm -0.75cm]{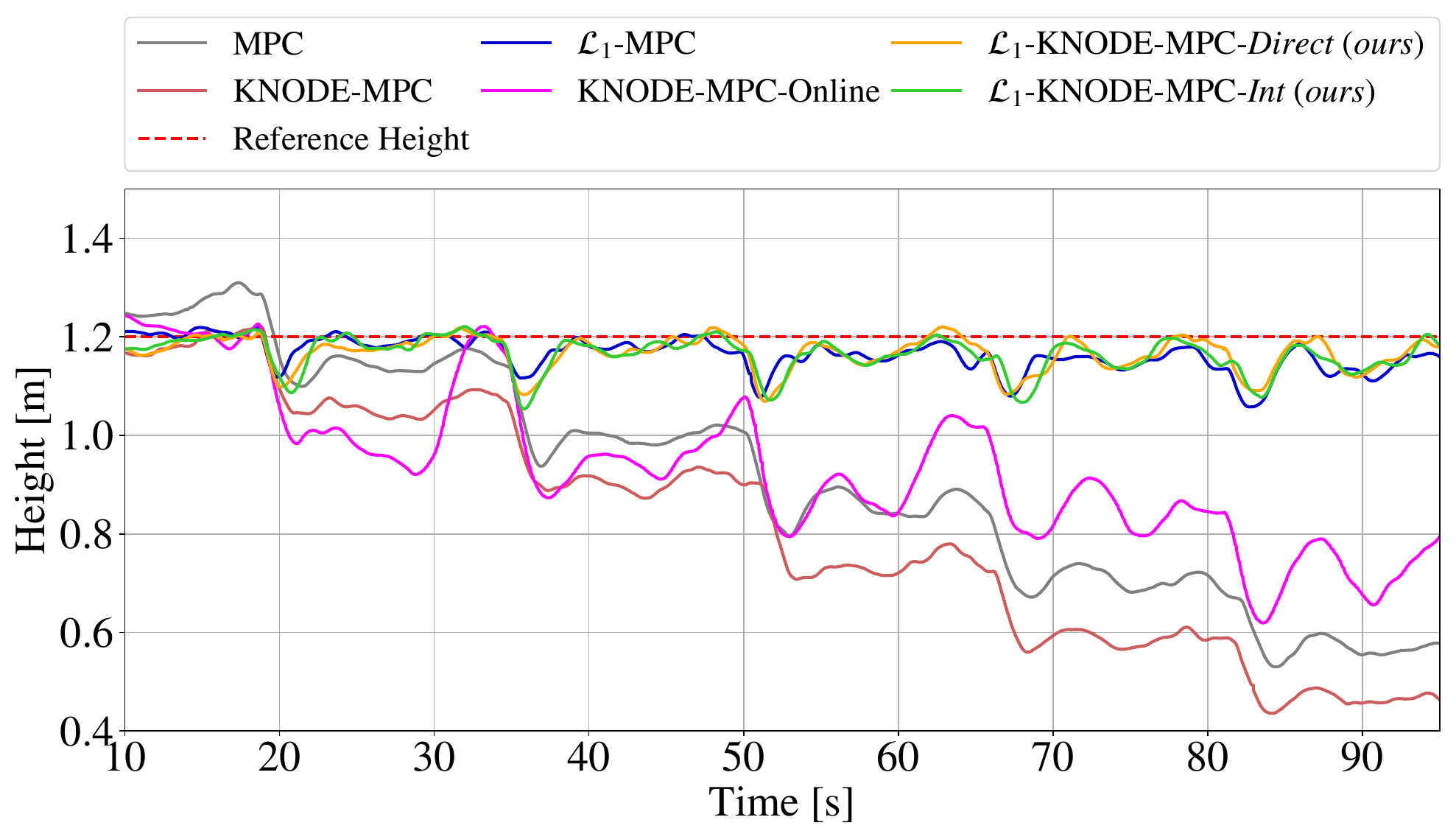}}
    \caption{\textbf{Height-time plots (ii):} Time histories of the quadrotor height for the benchmarks and proposed variants, under a more challenging test case in which payloads are added periodically. \vspace{-0.15cm}}
    \label{fig:testcase2_height}
\end{figure}

\vspace{-0.05cm}
To further investigate the adaptiveness of the framework, we consider a more challenging test case in which we attach slung payloads to the quadrotor periodically.
Starting from t=18.5s, a piece of payload is added every 15 seconds, until a total of 5 pieces are attached to the quadrotor system. These pieces have a total mass of approximately 4.9g. A depiction of this set of experiments can be found in the accompanying video. Fig. \ref{fig:testcase2_height} depicts the height-time plots of the quadrotor for this test case. Notably, the $\calL_1$ adaptation in the proposed framework allows the quadrotor to track the reference height closely.  Although the KNODE-MPC-Online method is able to alleviate the disturbances to some extent, the training of the NN takes time and hence, the height does not recover as quickly as compared to the proposed variants. While the $\calL_1$-MPC benchmark provides similar improvements in terms of height recovery, it has a larger RMSE in the $x$-$y$ plane, as compared to the proposed variants. Specifically, the $x$-$y$ RMSE for $\calL_1$-MPC is 22.0\% and 24.6\% higher than those of the two proposed variants. Overall, the average RMSE improvements brought forth by $\calL_1$-KNODE-MPC-\textit{Direct} and \textit{Int} are 44.9\% and 45.9\% above the benchmarks, demonstrating the effectiveness of the framework.

\vspace{-0.2cm}
\section{CONCLUSION}
\vspace{-0.1cm}
In this work, we propose $\calL_1$-KNODE-MPC, an adaptive learning-enhanced MPC framework that incorporates components from $\calL_1$ adaptive control into KNODE-MPC. We present two variants in our framework. The first variant fuses the $\calL_1$ adaptive module directly with KNODE-MPC at the level of control inputs, while the second variant integrates the uncertainty estimates from the $\calL_1$ adaptive module into the KNODE model, before applying it within KNODE-MPC. Through simulations and physical experiments, we show that the proposed
framework achieves significant performance improvements over a number of benchmarks. As future work, we plan to apply $\calL_1$-KNODE-MPC to other robotic applications to verify its generalizability.
\vspace{-0.15cm}
\section{Acknowledgments}
\vspace{-0.15cm}
The authors gratefully acknowledge the support of T. Z. Jiahao for providing the setup of the physical experiments.
\vspace{-0.2cm}
\bibliographystyle{IEEEtran} 
\bibliography{IEEEabrv,root} 

\addtolength{\textheight}{-12cm}   


\end{document}